\documentclass[runningheads]{llncs}

\usepackage{eccv}

\usepackage{eccvabbrv}
\usepackage[T1]{fontenc}
\usepackage{graphicx}
\usepackage{booktabs}
\usepackage{fontawesome5}
\usepackage[accsupp]{axessibility}  
\usepackage{subcaption}

\usepackage{hyperref}

\usepackage{orcidlink}
\usepackage{multirow}   
\usepackage{makecell}

\makeatletter
\def\@fnsymbol#1{\ensuremath{
  \ifcase#1\or *\or \dagger\or \ddagger\or
  \mathsection\or \mathparagraph\or \|\or **\or \dagger\dagger
  \or \ddagger\ddagger \else\@ctrerr\fi}}
\makeatother

\begin{document}



\title{Toward Physically Consistent Driving Video World Models under Challenging Trajectories}

\titlerunning{PhyGenesis: Physically Consistent Driving Video World Models}




\renewcommand{\thefootnote}{\fnsymbol{footnote}}




\author{
Jiawei Zhou\inst{1,2}\protect\footnotemark[1] \and
Zhenxin Zhu\inst{2}\protect\footnotemark[1] \and
Lingyi Du\inst{1}\protect\footnotemark[1] \and
Linye Lyu\inst{3} \and
Lijun Zhou\inst{2} \and
Zhanqian Wu\inst{2} \and
Hongcheng Luo\inst{2} \and
Zhuotao Tian\inst{4} \and
Bing Wang\inst{2} \and
Guang Chen\inst{2} \and
Hangjun Ye\inst{2} \and
Haiyang Sun\inst{2}\protect\footnotemark[2] \and
Yu Li\inst{1}\textsuperscript{\faEnvelope}
}

\authorrunning{J.~Zhou et al.}

\institute{
Zhejiang University \and
Xiaomi EV \and
The Hong Kong Polytechnic University \and
Shenzhen Loop Area Institute
}

\maketitle

\begingroup
\renewcommand{\thefootnote}{}
\footnotetext{$^{*}$ Equal Contribution. Email: \texttt{zhoujiawei6666@gmail.com}}
\footnotetext{$^{\dagger}$ Project Leader. Email: \texttt{sunhaiyang1@xiaomi.com}}
\footnotetext{\textsuperscript{\faEnvelope} Corresponding Author. Email: \texttt{yu.li.sallylee@gmail.com}}
\endgroup

\begin{figure}[htbp]
  \centering
  \includegraphics[width=\linewidth]{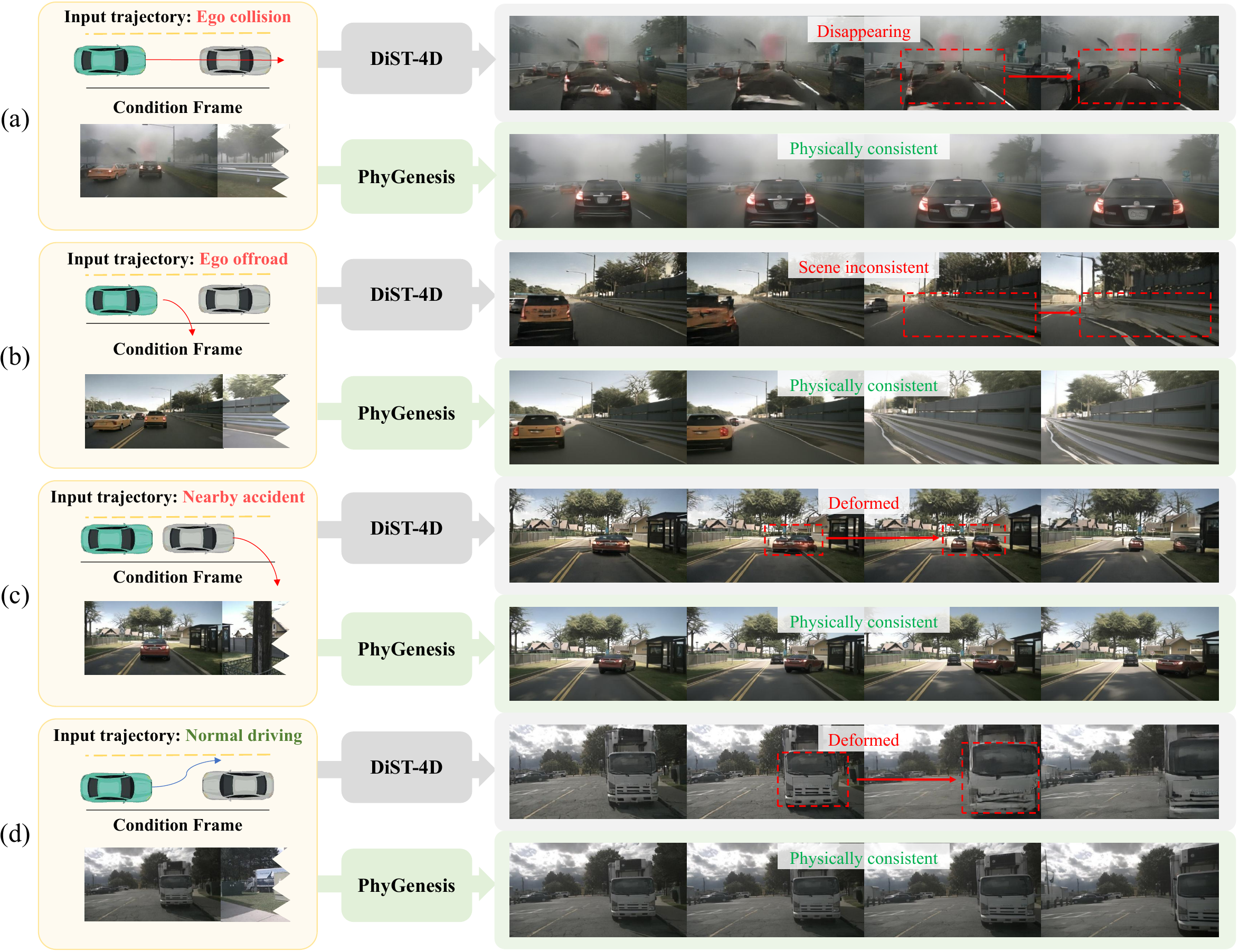}


  \caption{Qualitative comparison of video generation under diverse trajectory conditions (front view of the multi-view outputs is shown). Prior methods (e.g., DiST-4D) exhibit artifacts and geometric distortions under physically challenging trajectories, whereas \textbf{PhyGenesis} preserves physical consistency and high visual fidelity. Additional videos are provided in the supplementary material.}
    \label{fig:teaser}
\vskip -0.2in
\end{figure}

\begin{abstract}

 Video generation models have shown strong potential as world models for autonomous driving simulation.  However, existing approaches are primarily trained on real-world driving datasets, which mostly contain natural and safe driving scenarios. As a result, current models often fail when conditioned on challenging or counterfactual trajectories—such as imperfect trajectories generated by simulators or planning systems—producing videos with severe physical inconsistencies and artifacts.
To address this limitation, we propose \textbf{PhyGenesis}, a world model designed to generate driving videos with high visual fidelity and strong physical consistency. Our framework consists of two key components: (1) a \textit{physical condition generator} that transforms potentially invalid trajectory inputs into physically plausible conditions, and (2) a \textit{physics-enhanced video generator} that produces high-fidelity multi-view driving videos under these conditions.
To effectively train these components, we construct a large-scale, physics-rich heterogeneous dataset. Specifically, in addition to real-world driving videos, we generate diverse challenging driving scenarios using the CARLA simulator, from which we derive supervision signals that guide the model to learn physically grounded dynamics under extreme conditions. This challenging-trajectory learning strategy enables trajectory correction and promotes physically consistent video generation. 
Extensive experiments demonstrate that PhyGenesis consistently outperforms state-of-the-art methods, especially on challenging trajectories. Our project page is available at: \href{https://wm-research.github.io/PhyGenesis/}{https://wm-research.github.io/PhyGenesis/}.
\end{abstract}

\section{Introduction}
\label{sec:intro}

Video world models have recently emerged as a central paradigm for autonomous driving research~\cite{bruce2024genie,guo2025dist,ali2025world,gao2024enhance,hu2022model,hu2023gaia}, offering a scalable alternative to expensive real-world data collection and high-fidelity physical simulators.
Recent driving world models~\cite{gao2025magicdrive,zhao2025drivedreamer2,guo2025genesis,wen2024panacea,chen2026vilta,chen2025unimlvg,zeng2025rethinking} can synthesize high-fidelity multi-view future scenes while preserving controllability through structured conditions such as vehicle trajectories.
These capabilities have enabled a variety of downstream applications, including closed-loop evaluation in simulation~\cite{yang2025drivearena,yan2025drivingsphere}, high-risk scenario synthesis~\cite{zhou2025safemvdrive,xu2025challenger}, and integration with end-to-end planners for decision making and motion forecasting~\cite{zeng2025futuresightdrive,shi2025drivex,xia2025drivelaw,li2025recogdrive}.

Despite these advances, current driving world models struggle when deployed under \textit{challenging trajectory conditions} produced by trajectory simulators, planning systems, or user interactions. We identify two fundamental limitations of existing approaches.
\textbf{First, current models lack physical awareness of trajectory feasibility.} Trajectory conditions generated by simulators or planners can be imperfect and may violate fundamental physical constraints. However, existing models lack explicit physical reasoning and largely behave as condition-to-pixel translators. When forced to follow such physically inconsistent inputs, they often produce videos with severe rendering artifacts and structural failures.
\textbf{Second, current models lack physics-consistent generation capability.} Most existing approaches~\cite{gao2025magicdrive,gao2025magicdrive,guo2025genesis,wen2024panacea,chen2025unimlvg} are predominantly trained on real-world driving datasets dominated by safe and nominal behaviors. Consequently, they struggle to generate realistic dynamics in rare scenarios such as collisions or off-road departures—even when the trajectories themselves are physically feasible. 
Consequently, prior approaches (e.g., DiST-4D) often produce severe artifacts and physically inconsistent videos, as shown in Fig.~\ref{fig:teaser}.

In this work, we introduce \textbf{PhyGenesis}, a physics-aware driving world model designed to address both limitations. Our key insight is that physically consistent world modeling requires joint handling of trajectory feasibility and physics-consistent video generation. To this end, PhyGenesis introduces a novel module called the {\textit{Physical Condition Generator}}, which transforms arbitrary trajectory conditions into physically consistent ones by resolving potential physical conflicts. The rectified conditions are then fed into a {\textit{Physics-enhanced Video Generator}}, which synthesizes high-fidelity and physically consistent multi-view driving videos.
To support this learning process, we construct a heterogeneous training dataset that combines real-world driving data with a physically challenging dataset generated using the CARLA simulator~\cite{Dosovitskiy2017carla}. While real-world data provides abundant nominal driving behaviors, the CARLA-generated dataset introduces diverse extreme scenarios such as collisions and off-road departures. These events are uniquely informative, providing dense supervision for learning complex object–environment interactions—priors that are fundamentally scarce in routine real-world driving data.
With these designs, PhyGenesis substantially outperforms prior methods, particularly under challenging trajectory conditions.

Our main contributions are summarized as follows:
\begin{itemize}



 \item We propose \textbf{PhyGenesis}, a physics-aware driving world model for high-fidelity and physically consistent autonomous video generation. By explicitly handling both {trajectory feasibility} and {physics-consistent video generation}, PhyGenesis is the first framework capable of synthesizing physically consistent multi-view driving videos even when conditioned on initially physics-violating trajectory inputs.

\item We introduce a \textit{Physical Condition Generator} that converts arbitrary trajectory inputs into physically feasible 6-DoF vehicle motions. To enable this capability, we formulate a novel {counterfactual trajectory rectification training task} that equips the model with intrinsic physical priors for resolving physics-violating trajectories.



\item We develop a \textit{Physics-Enhanced Video Generator} and train these models on a heterogeneous physics-rich dataset combining real-world driving logs with physically extreme synthetic scenarios generated with the CARLA simulator. This hybrid training paradigm enables the model to learn complex object–environment interactions and significantly improves video generation under challenging trajectory conditions.



\end{itemize}

\section{Related Work}

\textbf{Nominal Driving World Models.} Driving video generation has progressed rapidly, with most methods conditioning on structured spatial priors for controllability. BEVGen~\cite{bevgen} encodes road and vehicle layouts via BEV maps but discards height information, limiting 3D representational capacity. BEVControl~\cite{bevcontrol} partially addresses this by introducing a height-lifting module to restore scene geometry. MagicDrive~\cite{gao2023magicdrive} further advances 3D-aware generation through geometric constraints and cross-view attention, while MagicDrive-V2~\cite{gao2025magicdrive} adopts Diffusion Transformers for higher-resolution, temporally coherent synthesis. DriveDreamer~\cite{wang2024drivedreamer} introduces hybrid Gaussians for temporally consistent complex maneuver rendering. DiST-4D~\cite{guo2025dist}  and WorldSplat~\cite{zhu2025worldsplat} incorporate metric depth to lift generated videos into 4D scene representations for novel viewpoint synthesis. In the multimodal direction, Genesis~\cite{guo2025genesis} and UniScene~\cite{li2024uniscene} target joint LiDAR--RGB generation via sequential DiT and occupancy-centric voxel representations, respectively. 
While these methods achieve high fidelity under routine driving, their reliance on nominal datasets limits robustness to challenging and/or physics-violating trajectory inputs.

\setlength{\parskip}{1pt}
\noindent\textbf{High-risk Driving Video Generation.} Generating high-risk driving scenarios has attracted growing attention. 
Early efforts such as AVD2~\cite{li2025avd2}, DrivingGen~\cite{guo2024drivinggen}, and Ctrl-Crash~\cite{gosselin2025ctrlcrash} synthesize accident scenarios from single-view dashcam footage; however, their single-view, low-quality data making it difficult for models trained on these datasets to transfer to high-fidelity, multi-view simulators.
More recent methods, SafeMVDrive~\cite{zhou2025safemvdrive} and Challenger~\cite{xu2025challenger}, combine trajectory simulators with multi-view video generators to produce safety-critical videos. Nevertheless, their video generators are trained exclusively on nominal data, so the resulting quality remains limited and the generated scenes cannot depict physical interactions such as collisions.


In summary, existing driving world models handle either nominal scenarios or high-risk synthesis, but rarely both. \textbf{PhyGenesis} bridges this gap through a Physical Condition Generator and a Physics-Enhanced Video Generator trained on both real-world and simulation-derived extreme data.

\section{Method}
\label{sec:method}

\begin{figure}[tb]
  \centering
  \includegraphics[width=\linewidth]{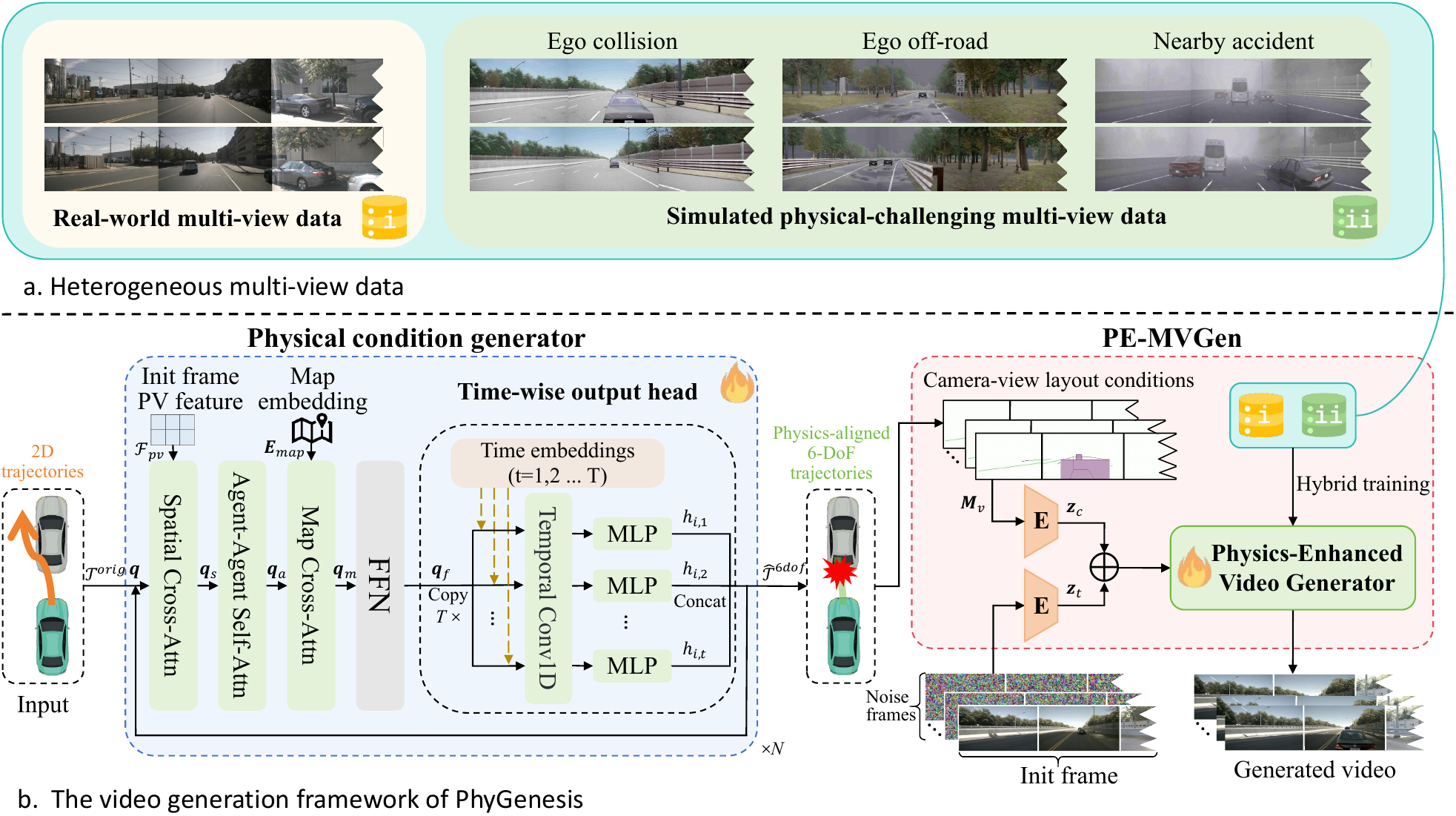}
  \caption{ Overview of \textbf{PhyGenesis}. (a) Our heterogeneous multi-view dataset consists of both real-world driving data and simulated data that emphasizes physically challenging scenarios, including ego-vehicle collisions and roadway departures, among others. (b)~In PhyGenesis, the \textit{physical condition generator} first rectifies arbitrary 2D trajectories—potentially counterfactual or physics-violating—into physically plausible 6-DoF motions. The rectified trajectories are then projected into camera-view layout conditions and fed into a \textit{physics-enhanced video generator}, co-trained on the heterogeneous dataset, to synthesize high-fidelity, physically consistent multi-view videos.}
  \vskip -0.1in
  \label{fig:framework}
\end{figure}

\subsection{Overview of PhyGenesis}
\label{sec:overview}

The overview of our \textbf{PhyGenesis} framework is illustrated in Figure \ref{fig:framework}. As shown in (a), our framework is trained on a heterogeneous multi-view dataset (Section \ref{Heterogeneous Data}), which enables the model to learn both high visual fidelity and physical consistency, even under challenging scenarios.
Given trajectory inputs, the \textit{Physical Condition Generator} (Section \ref{Physical condition generator}) first rectifies potentially invalid trajectories into physically consistent 6-DoF vehicle motions.
These rectified conditions are then passed to the \textit{Physics-Enhanced Multi-view Video Generator} (PE-MVGen) (Section \ref{Physics-Enhanced Multi-view Video Generator}), which synthesizes  multi-view video sequences.
This unified pipeline enables high quality video generation even when the input trajectories violate physical constraints.

Specifically, our system takes as inputs the initial multi-view images $\mathcal{I}_0$, static map $\mathcal{M}$, and a set of future trajectories $\mathcal{T}^{orig}$ for all $N$ agents, \textit{i.e.}, cars. We define $\mathcal{T}^{orig}=\{\mathcal{T}^{orig}_i\}_{i=1}^{N}$, where $\mathcal{T}^{orig}_i=\{\mathcal{T}^{orig}_{i,t}\}_{t=1}^{T}$ and $\mathcal{T}^{orig}_{i,t}=(x_{i,t},y_{i,t})$ specifies the 2D location of agent $i$ at time $t$. This 2D trajectory representation $\mathcal{T}^{orig}$ aligns with the standard output format of mainstream trajectory simulators and end-to-end autonomous driving planners. Crucially, these trajectories can be physics-violating (e.g., containing overlapping paths that would cause object penetration). Given such potentially flawed inputs, our goal is to synthesize a high-fidelity, multi-view video sequence $\mathcal{V}_{1:T}$ that faithfully reflects the intended driving behaviors while adhering to real-world physical constraints. Next, we detail each part of our design.

\subsection{Heterogeneous Multi-view Data}
\label{Heterogeneous Data}

\textbf{Real-World Multi-view Data.} Following the recent multi-view driving world models \cite{gao2025magicdrive,zhao2025drivedreamer2}, we utilize nominal real-world driving logs from the \textit{nuScenes} dataset \cite{caesar2020nuscenes} to establish a foundational understanding of complex urban environments. However, these data are heavily biased towards safe driving behaviors and inherently lack complex physical interactions (\eg collisions, off-road driving). This data deficiency causes generative models to produce artifacts and physically inconsistent motion under physically challenging trajectories.

\setlength{\parskip}{1pt}
\noindent \textbf{Simulated Physically-challenging Multi-view Data.} To empower a world model with robust physical understanding and the capability to generate videos under physically challenging interactions, training data that explicitly contains such events is essential. Since collecting safety-critical real-world data is impractical, modern driving simulators (e.g., CARLA ~\cite{Dosovitskiy2017carla}) provide high-fidelity physics engines and controllable environmental variations. Prior work like ReSim \cite{yang2025resim} has attempted to incorporate synthetic data into world-model training to mitigate the limited coverage of real-world distributions. Their synthetic data is limited to a single view with only ego-trajectory annotations, which makes it difficult to train models that control multiple agents. Moreover, their data collection is not explicitly focused on physically challenging events, providing limited supervision for strengthening a model's physical priors. 

\begin{figure}[tb]
  \centering
  \includegraphics[width=0.6\linewidth]{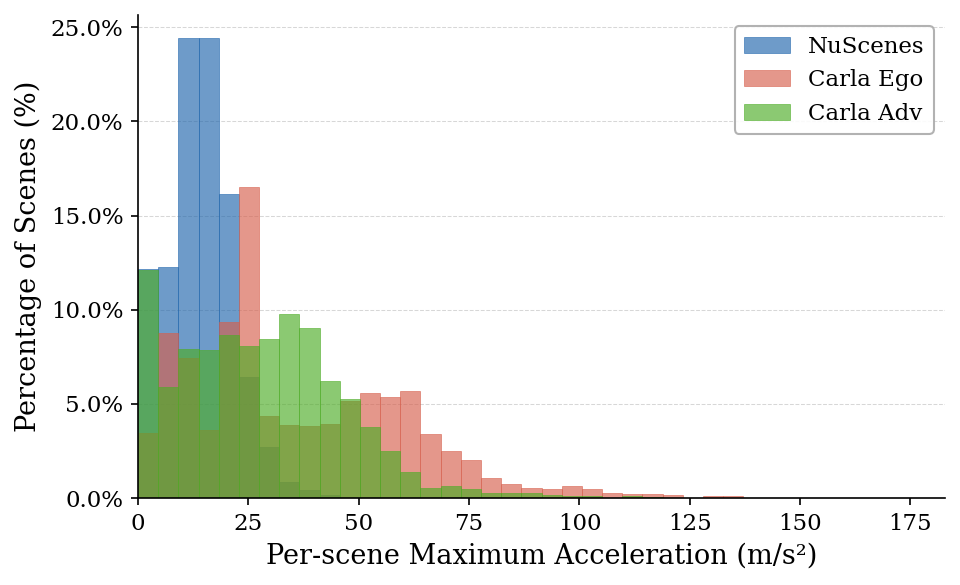}
  \vskip -0.1in
  \caption{ 
Distributions of maximum ego-vehicle acceleration for \textit{nuScenes}, \textit{CARLA Ego}, and \textit{CARLA ADV}. The simulated CARLA datasets show a clear shift toward higher accelerations, indicating more aggressive dynamics and physically challenging events compared with the predominantly nominal driving behaviors in \textit{nuScenes}.
  }
  \label{fig:compare_ego_accel}
  \vskip -0.1in
\end{figure}

To fill this gap, we leverage the CARLA simulator to build a large-scale multi-view synthetic dataset focused on physically challenging scenarios. We follow the Bench2Drive routing setup~\cite{jia2024bench2drive} to cover diverse scenes, weather, and traffic events. Based on this foundation, we curate two subsets: \textit{CARLA Ego}, capturing interactions between the ego vehicle and the environment or surrounding agents, and \textit{CARLA Adv}, capturing interactions centered on a nearby non-ego agent. During collection, we perturb the route and target speed of the ego vehicle (or the Adv agent) to induce collisions, off-road departures, and abrupt maneuvers (Detailed in supplementary material). This results in substantially more aggressive dynamics than \textit{nuScenes}, as reflected by the shifted maximum ego-acceleration distribution in Figure~\ref{fig:compare_ego_accel}.


We equip the simulation with a sensor suite rigorously aligned with the \textit{nuScenes} configuration, comprising 1 LiDAR, 6 surround-view cameras ($900 \times 1600$ resolution), 5 radars, and 1 IMU/GNSS unit. Crucially, to accurately capture physical anomalies, we additionally integrate a collision sensor and high-definition (HD) map metadata, allowing us to precisely record the exact timestamps of impacts and off-road moments. The data is recorded at 12Hz and is meticulously annotated and organized into a format identical to the \textit{nuScenes} dataset to ensure seamless downstream training.

\setlength{\parskip}{1pt}
\noindent \textbf{Heterogeneous Dataset Construction.} In total, we simulated approximately 31 hours of driving data. The \textit{CARLA-Adv} subset contains 15.5 hours with 760K annotated bounding boxes, while the \textit{CARLA-Ego} subset comprises 15.2 hours with 830K boxes. Utilizing explicit collision sensor signals and map metadata, we design a rule-based filtering mechanism to precisely localize timestamps of physical interactions, extracting 9.7 hours of highly physically-challenging video clips. Finally, we combine these 9.7 hours of simulated clips with 4.6 hours of real-world data to construct our heterogeneous dataset.

\subsection{Physical Condition Generator}
\label{Physical condition generator}
Since rendering videos from physics-violating 2D layouts often causes severe artifacts (e.g., distortion or melting), we introduce a \textit{Physical Condition Generator} as the first stage of our framework. Its dynamically rectify the possibly physical-violating $T$-frame 2D trajectories $\mathcal{T}^{orig}$ into a physically plausible 6-DoF trajectory sequence $\hat{\mathcal{T}}^{\text{6dof}}$. The transition to 6-DoF (including $x, y, z$, pitch, yaw, roll) is crucial, as extreme physical interactions often induce drastic variations in the vertical and rotational axes that 2D coordinates cannot capture.

\begin{figure}[tb]
  \centering
  \includegraphics[width=\linewidth]{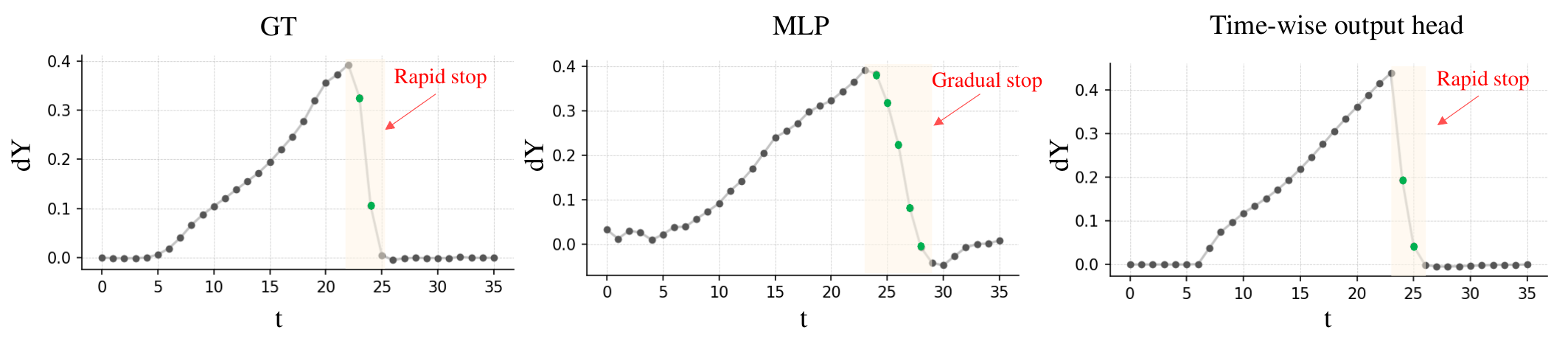}
  \vskip -0.1in
  \caption{ 
  Comparison of MLP and time-wise output head in simulating collision dynamics. The MLP shows a gradual velocity decrease after the collision, while the GT and time-wise head show an instantaneous drop to zero, producing more realistic dynamics.
  }
  \vskip -0.1in
  \label{fig:6dof-head}
\end{figure}

\noindent \textbf{Architecture.} The architecture of the physical model is illustrated in the left of Figure \ref{fig:framework} (b). For a given scenario, the input trajectories $\mathcal{T}^{orig}$ are encoded via sine--cosine positional encoding followed by an MLP encoder into agent tokens $\mathbf{q} \in \mathbb{R}^{N \times D}$, where $N$ denotes the number of agents and $D$ is the token dimension. To ensure that these agents interact reasonably with the visual environment, we first apply deformable spatial cross-attention to interact with the multi-view Perspective View (PV) features $\mathcal{F}_{pv}$ based on their trajectory coordinates. This operation yields the spatially grounded queries $\mathbf{q}_{s}$:
\begin{equation}
    \mathbf{q}_{s} = \text{SpatialCrossAttn}(\mathbf{q}, \mathcal{F}_{pv})
\end{equation}
Subsequently, an agent-agent self-attention layer is introduced to enable the tokens to perceive the positional and kinematic states of surrounding vehicles, which is the key design for resolving overlapping and penetration conflicts:
\begin{equation}
    \mathbf{q}_{a} = \text{AgentSelfAttn}(\mathbf{q}_{s})
\end{equation}
Furthermore, a map cross-attention layer integrates vectorized map embeddings $\mathbf{E}_{map}$ for better off-road awareness:
\begin{equation}
    \mathbf{q}_{m} = \text{MapCrossAttn}(\mathbf{q}_{a}, \mathbf{E}_{map})
\end{equation}
Finally, a Feed-Forward Network is applied to non-linearly transform the aggregated features, yielding the fully refined queries $\mathbf{q}_{f}$ before trajectory prediction:
\begin{equation}
    \mathbf{q}_{f} = \text{FFN}(\mathbf{q}_{m})
\end{equation}

Following the above layers, the refined queries need to be projected into the final trajectories. Traditional MLPs typically smooth out trajectory outputs, failing to capture the sudden, high-frequency dynamic impulses indicative of a collision. To address this, we specifically design a \textit{Time-Wise Output Head} as the final prediction module. For the $i$-th refined agent token $\mathbf{q}_f[i]$, we expand it across the $T$ future steps and concatenate it with a step-specific learnable temporal embedding $\mathbf{E}_{time}(t)$. The concatenated feature is then processed by a Temporal Convolutional Network (TCN) to capture local inter-step dynamic variations, before being projected by an MLP to output the exact 6-DoF state:
\begin{equation}
    \mathbf{h}_{i,t} = \text{TCN} \left( \text{Proj}(\mathbf{q}_f[i] \parallel \mathbf{E}_{time}(t)) \right)
\end{equation}
\begin{equation}
    \hat{\mathcal{T}}_{i,t}^{6dof} = \text{MLP}(\mathbf{h}_{i,t}) \in \mathbb{R}^6
\end{equation}
As shown in Figure~\ref{fig:6dof-head}, unlike standard regression heads that produce sluggish responses, this time-wise formulation paired with step-specific time embeddings accurately captures the abrupt physical changes at the physical impact moment.


\setlength{\parskip}{1pt}
\noindent \textbf{Training Pair Construction.}  To equip the Physical Condition Generator with the ability to rectify physics-violating trajectory conditions, we construct paired training data that maps \emph{physics-violating} trajectory inputs to physically feasible targets. To achieve this, we propose a systematic counterfactual trajectory corruption strategy. Specifically, for a collision clip in our simulated  dataset, we keep the original trajectory logs before the collision. For the post-collision frames, we intentionally corrupt the trajectories of all agents by extending their path with the same velocity before the collision, which synthesizes penetration-style counterfactual trajectory conditions. The ground-truth simulation logs---which capture the actual collision dynamics---serve as the supervision target for correction. In addition, to avoid distorting natural driving conditions, we also include real-world nominal trajectory-condition pairs from nuScenes without applying counterfactual corruption, ensuring the model preserves realistic inputs while learning to rectify physically invalid ones.



\setlength{\parskip}{1pt}
\noindent \textbf{Optimization.} The Physical Model is optimized using a weighted distance loss between the predicted 6-DoF trajectories and the ground truth:
\begin{equation}
    \mathcal{T}_{phy} = \frac{1}{N \times T} \sum_{i=1}^{N} \sum_{t=1}^{T} W_{i,t} \left\| \hat{\mathcal{T}}_{i,t}^{6dof} - \mathcal{T}_{i,t}^{gt} \right\|_1
\end{equation}
To focus on critical physical moments, we define $W_{i,t}$ with two scalars: an event-window weight $\lambda_{\text{event}}$ that increases the loss within a temporal window around collision/off-road timesteps, and a physical-agent weight $\lambda_{\text{agent}}$ that further amplifies the loss for agents involved in the interaction.

\subsection{Physics-Enhanced Multi-view Video Generator}
\label{Physics-Enhanced Multi-view Video Generator}

The second stage of our framework is the Physics-Enhanced Multi-View Video Generator (PE-MVGen). Built upon Wan2.1 \cite{wan2025wan}, a high-capacity diffusion transformer (DiT) originally conditioned on images and text, we adapt it into a controllable, multi-view generator explicitly designed for the autonomous driving domain. Crucially, this generator is endowed with deep physical awareness through a specialized heterogeneous co-training strategy.

\setlength{\parskip}{1pt}
\noindent \textbf{Multi-View~\& Layout Conditioning.} We first encode the input multi-view clips into latents $\mathbf{z}\in\mathbb{R}^{V\times T\times C\times h\times w}$ using a pre-trained 3D VAE~\cite{wan2025wan}, where $V$ represents the number of views. To enable multi-view modeling without introducing additional parameters \cite{gao2025magicdrive,wen2024panacea}, we reshape the input into $ T\times C\times h \times ( V \cdot w) $, concatenating the view dimension into the spatial axis so that the same self-attention can capture cross-view dependencies. Furthermore, to explicitly condition generation on structural layouts, we project the future $T$-frame 3D agent boxes and map polylines onto each camera view using the calibrated intrinsics $\mathbf{K}_v$ and extrinsics $\mathbf{E}_v$. The resulting view-specific control images $\mathbf{M}_v$ are encoded by the VAE encoder to get $\mathbf{z}_c$ , reshaped to match the video latents, concatenated along the channel dimension with the noisy latent input $\mathbf{z}_t$, and finally processed by a patch embedder before entering the DiT.

\setlength{\parskip}{1pt}
\noindent \textbf{Data-Driven Physical Enhancement.} Current world models fail in physically challenging scenarios because their training distributions lack physical interactions. To solve this, we train PE-MVGen on our heterogeneous dataset, maintaining a balanced 1:1 ratio between nominal real-world logs and simulated physically challenging data. Crucially, we do not use counterfactual trajectories at this stage; the generator is supervised with ground-truth physical trajectories, decoupling physical correction from rendering. As demonstrated in Figure \ref{fig:with_physical_data}, co-training with this physics-rich data significantly enhances the model's physical understanding, allowing its generative capabilities to robustly generalize to physically challenging scenarios in real-world. \cite{ge2025unraveling,fang2025rebot}

\setlength{\parskip}{1pt}
\noindent \textbf{Training Objective.} Following Wan 2.1, PE-MVGen is optimized via Rectified Flows \cite{esser2024scaling}, which ensures stable training through ordinary differential equations (ODEs). During training, given a clean video latent $\mathbf{z}_1$, a random noise $\mathbf{z}_0 \sim \mathcal{N}(\mathbf{0}, \mathbf{I})$, and a timestep $t \in [0,1]$ sampled from a logit-normal distribution, the intermediate noisy latent $\mathbf{z}_t$ is defined as a linear interpolation:
\begin{equation}
    \mathbf{z}_t = t\mathbf{z}_1 + (1 - t)\mathbf{z}_0
\end{equation}
The ground-truth velocity vector is defined as $\mathbf{v}_t = \mathbf{z}_1 - \mathbf{z}_0$. The DiT model, parameterized by $\theta$, is trained to predict this velocity $u_\theta$. The flow-matching objective is formulated as the Mean Squared Error (MSE):
\begin{equation}
    \mathcal{L}_{FM} = \mathbb{E}_{\mathbf{z}_0, \mathbf{z}_1, t} \left\| u_\theta(\mathbf{z}_t, t, \mathbf{c}_{init}, \mathbf{c}_{text}, \mathbf{c}_{layout}) - \mathbf{v}_t \right\|_2^2
\end{equation}
where the predicted velocity $u_\theta$ is conditioned on three key components: $\mathbf{c}_{init}$ denotes the latent features of a single initial context frame, $\mathbf{c}_{text}$ represents the scene caption, and $\mathbf{c}_{layout}$ encodes the future multi-view layout images.

\setlength{\parskip}{1pt}
\noindent \textbf{Curriculum Co-Training Strategy.} We adopt a highly efficient two-stage curriculum: pre-training at a lower resolution ($224 \times 400$) to quickly learn multi-view geometry and physically challenging layout mappings, followed by high-resolution ($448 \times 800$) fine-tuning to ensure visual fidelity.

\section{Experiment}
\subsection{Experimental Setup}
\textbf{Datasets:} We use the heterogeneous multi-view dataset introduced in Section~\ref{Heterogeneous Data}, including \textit{CARLA Ego}, \textit{CARLA ADV}, and \textit{nuScenes}, for training and evaluation. During training, we adopt balanced sampling to keep the simulated-to-real clip ratio close to 1:1. For video generation evaluation, we randomly sample 150 clips per test split and generate videos for all sampled clips. Since the compared baselines are primarily trained on \textit{nuScenes}, we translate the CARLA clips into the \textit{nuScenes} visual style using a learned style-transfer model (details in the supplementary material) for fair evaluation.

\setlength{\parskip}{1pt}
\noindent \textbf{Evaluation Metrics:}  
We evaluate generated videos along three dimensions: visual quality, physical plausibility, and controllability. \textbf{(1):} Following prior work, we use \textbf{FID} and \textbf{FVD} to measure the realism. 
 \textbf{(2):} For physical plausibility, we adopt WorldModelBench \cite{li2025worldmodelbench}, a benchmark that can effectively measure the effectiveness of autonomous driving video generation using human preference-aligned VLM-based judges.  We report \textbf{PHY} as the average of four WorldModelBench metrics: Mass (objects do not deform irregularly), Impenetrability (objects do not unnaturally pass through each other), Frame-wise Quality (no unappealing frames or low-quality content), and Temporal Quality (no temporally inconsistent scenes and objects).  In addition, we report a human preference rate (\textbf{Pref.}), computed as the percentage of pairwise comparisons in which a method is preferred by human annotators. Detailed questionnaire and settings are provided in the supplementary material. \textbf{(3):} For controllability, we measure how well the generated videos follow the input trajectory conditions. Specifically, we compute the Controllability Error (\textbf{CtrlErr}) between the trajectories extracted from generated videos and the ground-truth trajectories, defined as the geometric mean of Rotation Error and Translation Error, following prior work\cite{he2024cameractrl,duan2025worldscore}. Camera pose sequences are extracted using ViPE\cite{huang2025vipe}. 

\setlength{\parskip}{1pt}
\noindent \textbf{Baseline:}  
We compare our method with three baselines:  UniMLVG \cite{chen2025unimlvg}, MagicDrive-V2 \cite{gao2025magicdrive}, and DiST-4D \cite{guo2025dist}, evaluating their performance on our dataset. For UniMLVG, MagicDrive-V2, and our framework, the input includes the initial RGB frame, the scene captions, and future layout information. For DiST-4D, it additionally requires the initial-frame depth map alongside the initial RGB frame, providing extra geometric cues. Implementation details of these baselines are provided in the supplementary material.

\setlength{\parskip}{1pt}
\noindent \textbf{Implementation Details:}  
The training of our Physical Condition Generator conducted on 12Hz data with $T=36$. Perception-view (PV) features are extracted using a ResNet50 \cite{he2016deep} initialized from \cite{sun2025sparsedrive} and trained with a learning rate of $9 \times 10^{-5}$. The main network is trained with a learning rate of $9 \times 10^{-4}$, and the batch size is set to 256. We set $N=2$ and for the loss weight $W_{i,t}$  in Section \ref{Physical condition generator}, we set $\lambda_{\text{event}}=10$ and $\lambda_{\text{agent}}=5$; an ablation over these hyperparameters is provided in the supplementary material.


Our video generation model is initialized from the pre-trained WAN2.1 weights and optimized with AdamW. We adopt a two-stage schedule: Stage~1 trains at $224\times 400$ for 2{,}850 steps with a learning rate of $5\times 10^{-5}$ and a global batch size of 480, followed by Stage~2 at $448\times 800$ for 350 steps with a learning rate of $1\times 10^{-4}$ and a global batch size of 240. Training is implemented on 48 NVIDIA H20 GPUs. The model generates 33-frame videos at 12\,Hz.

\subsection{Performance of PhyGenesis}\label{exp:complete framework}

\vskip -0.1in
\begin{figure}[tb]
  \centering
  \includegraphics[width=\linewidth]{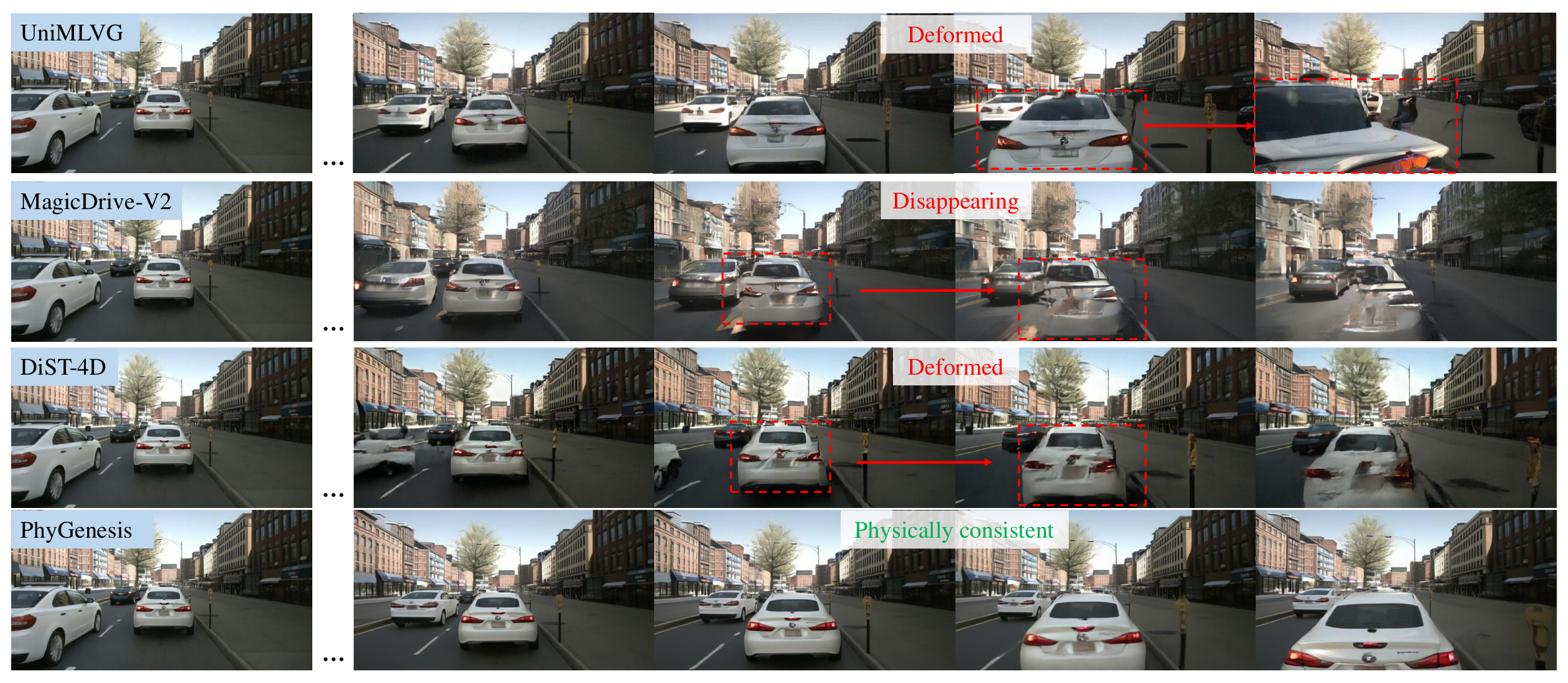}
  \vskip -0.1in
  \caption{ 
  Qualitative comparison with different baselines. Our method maintains the best physical consistency and produces the best visual quality.
  }
  \vskip -0.1in
  \label{fig:basline_2d_2}
\end{figure}

\begin{figure}[tb]
  \centering
  \includegraphics[width=\linewidth]{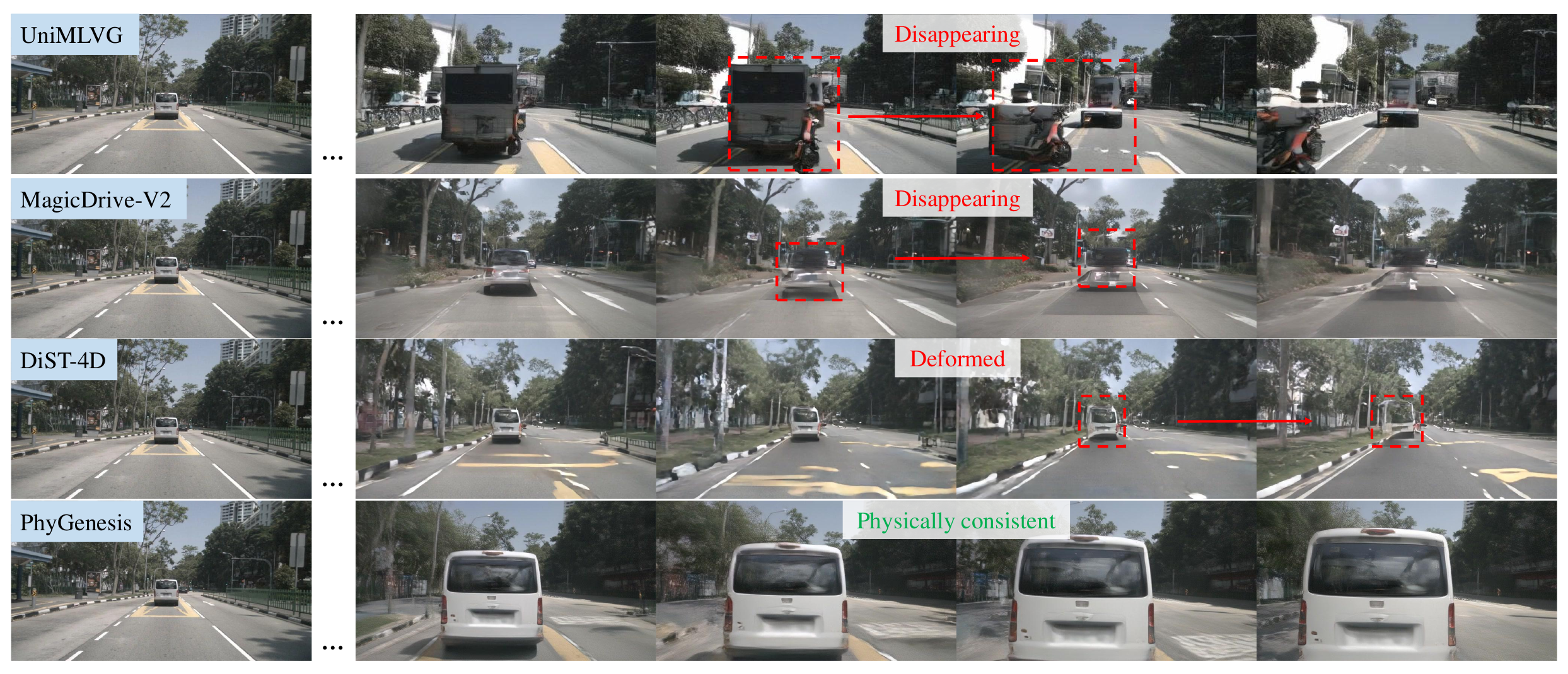}
  \vskip -0.1in
  \caption{ 
  Qualitative comparison on the \textit{nuScenes} stress test under out-of-distribution (OOD), physics-violating trajectory inputs, with the first-frame condition kept unchanged. 
 Our method outperforms baselines with
  respect to physical consistency.
  }
  \label{fig:nuscenes-collision}
  \vskip -0.1in
\end{figure}


\begin{table}[t]

\caption{Comparison with baselines under 2D trajectory conditions: \textit{nuScenes} uses nominal trajectories, while \textit{CARLA Ego} and \textit{CARLA ADV} use physics-violating trajectories. We report visual quality (FID/FVD), physical consistency (PHY), human preference (Pref.). PhyGenesis achieves the best results across all datasets.}
\vskip -0.15in
\label{tab:performance_complete}
\begin{center}
\begin{small}
\begin{sc}
\resizebox{\columnwidth}{!}{%
\begin{tabular}{l c cccc cccc cccc}
\toprule
\multirow{2}*{\textbf{Methods}}&
\multicolumn{4}{c}{nuScenes} & \multicolumn{4}{c}{CARLA Ego} & \multicolumn{4}{c}{CARLA ADV} \\
\cmidrule(lr){2-5} \cmidrule(lr){6-9} \cmidrule(lr){10-13}
& FID$\downarrow$ & FVD$\downarrow$ & PHY$\uparrow$  & Pref. $\uparrow$
  & FID$\downarrow$ & FVD$\downarrow$ & PHY$\uparrow$ & Pref. $\uparrow$
  & FID$\downarrow$ & FVD$\downarrow$ & PHY$\uparrow$ & Pref. $\uparrow$ \\
\midrule
UniMLVG  \cite{chen2025unimlvg}             & 17.59   & 129.69    & 0.93 & 0.05 & 34.50   & 260.21    & 0.55 & 0.13 & 30.32   & 274.94    & 0.70 & 0.19 \\
MagicDrive-V2 \cite{gao2025magicdrive}        & 13.40 & 91.10 & 0.92& 0.16 & 32.19 &  207.64 & 0.60 & 0.06& 32.82 & 222.55 & 0.66 & 0.10 \\
DiST-4D \cite{guo2025dist}        & 10.49 & 46.95 & 0.86&  0.13 & 19.84 & 197.57 & 0.39 & 0.10& 16.07 & 128.88 & 0.56 & 0.05\\
PhyGenesis        & \textbf{10.24} & \textbf{40.41}  & \textbf{0.97} & \textbf{0.67} & \textbf{11.03} & \textbf{72.48}  & \textbf{0.71}& \textbf{0.71} & \textbf{9.28} & \textbf{77.83} & \textbf{0.87}& \textbf{0.66} \\
\bottomrule
\end{tabular}}
\end{sc}
\end{small}
\end{center}
\vskip -0.1in
\end{table}

In this section, we evaluate the effectiveness of \textbf{PhyGenesis}. For \textit{nuScenes}, we use the ground-truth $(x,y)$ trajectories as the input trajectory condition $\mathcal{T}^{orig}$. For \textit{CARLA Ego} and \textit{CARLA ADV}, to assess generation under physically-violating trajectory inputs, we construct counterfactual, physics-violating conditions following Section~\ref{Physical condition generator}: we keep the pre-collision trajectories unchanged and extrapolate the post-collision motion using each agent's pre-collision velocity. For baseline methods, we additionally compute yaw angles from $(\Delta x,\Delta y)$ to simulate rotation for a fair comparison~\cite{liao2025diffusiondrive}. As shown in Table~\ref{tab:performance_complete}, our framework achieves the best physical scores across all datasets and attains the best visual quality, with the largest gains on the physically challenging CARLA sets, , where physics-violating inputs often cause deformation and penetration in prior methods (Figure~\ref{fig:basline_2d_2}).

To evaluate video generation under out-of-distribution, physically challenging trajectories in real-world scenes, we additionally conduct a \textit{nuScenes} stress test. Specifically, we create physics-violating trajectory conditions in the \textit{nuScenes} test set by scaling up the ego vehicle speed and retaining only collision cases, while keeping the first-frame condition unchanged. Qualitative and quantitative results in Figure~\ref{fig:nuscenes-collision} and Figure~\ref{fig:nuscenes_violation_bar} show that \textbf{PhyGenesis} remains more physically consistent under these corrupted conditions.

\subsection{Performance of Physics-enhanced Multi-view Video Generator}

\begin{table}[t]
\vskip -0.1in
\caption{Comparison of video generators of different baselines under GT trajectory conditions. We report visual realism (FID/FVD), physical consistency (Phy), and controllability error (CtrlErr); best results are in \textbf{bold}.}
\vskip -0.1in
\label{tab:performance_video_generator}
\begin{center}
\begin{small}
\begin{sc}
\resizebox{\columnwidth}{!}{
\begin{tabular}{lcccccccccccccccc}
\toprule
\multirow{2}*{\textbf{Methods}} & \multicolumn{4}{c}{\textit{nuScenes}} & \multicolumn{4}{c}{\textit{CARLA Ego}} & \multicolumn{4}{c}{\textit{CARLA ADV}} \\
\cmidrule(lr){2-5} \cmidrule(lr){6-9} \cmidrule(lr){10-13}
& FID $\downarrow$ & FVD$\downarrow$  & PHY $\uparrow$  & CtrlErr $\downarrow$
& FID $\downarrow$  & FVD $\downarrow$  & PHY $\uparrow$ & CtrlErr $\downarrow$
& FID $\downarrow$  & FVD $\downarrow$  & PHY $\uparrow$ & CtrlErr $\downarrow$ \\
\midrule
UniMLVG \cite{chen2025unimlvg}       &  16.63   & 113.23    &  0.92 & 0.35  & 34.03 & 240.78  & 0.56    & 1.32  &  30.06 & 264.59 & 0.75   & 0.66     \\
MagicDrive-V2 \cite{gao2025magicdrive} & 13.17 & 88.86 & 0.93 & 0.29  & 32.92 & 181.26 & 0.58 & 1.26 & 33.62 & 214.52 & 0.65 & 0.75 \\
DiST-4D \cite{guo2025dist}     & 10.48 & 45.24 & 0.84 & 0.28 & 19.94 & 133.10 & 0.38 & 1.19 & 16.12 & 105.70 & 0.50 &  0.57\\
PhyGenesis & \textbf{10.20} & \textbf{31.14}  & \textbf{0.97} & \textbf{0.25}& \textbf{10.98} & \textbf{57.02} & \textbf{0.69} & \textbf{0.85}  & \textbf{9.07} & \textbf{59.44} & \textbf{0.83} & \textbf{0.37} \\
\bottomrule
\end{tabular}
}
\vskip -0.1in
\end{sc}
\end{small}
\end{center}
\end{table}


In this section, we evaluate the Physics-Enhanced Video Generator. We condition all methods on \emph{ground-truth} trajectories (i.e., no counterfactual or physics-violating inputs) and report results on \textit{nuScenes}, \textit{CARLA Ego}, and \textit{CARLA Adv} in Table~\ref{tab:performance_video_generator}. Although the inputs are physically feasible, prior baselines are trained mostly on nominal logs and rarely see physics-rich interactions (e.g., collisions and off-road events), so they still struggle to render these dynamics, resulting in worse visual quality, physical consistency, and controllability. In contrast, our PE-MVGen is co-trained with physics-rich data, better capturing complex interactions and improving both generation quality and condition following.

\subsection{Performance of Physical Condition Generator}



\begin{table}[tb]
\centering

\begin{minipage}[t]{0.49\linewidth}
\centering
\captionof{figure}{Human preference and Physical score of baseline generations on \textit{nuScenes} under physics-violating trajectories.}
\label{fig:nuscenes_violation_bar}
\vspace{-0.9em}
\includegraphics[width=\linewidth]{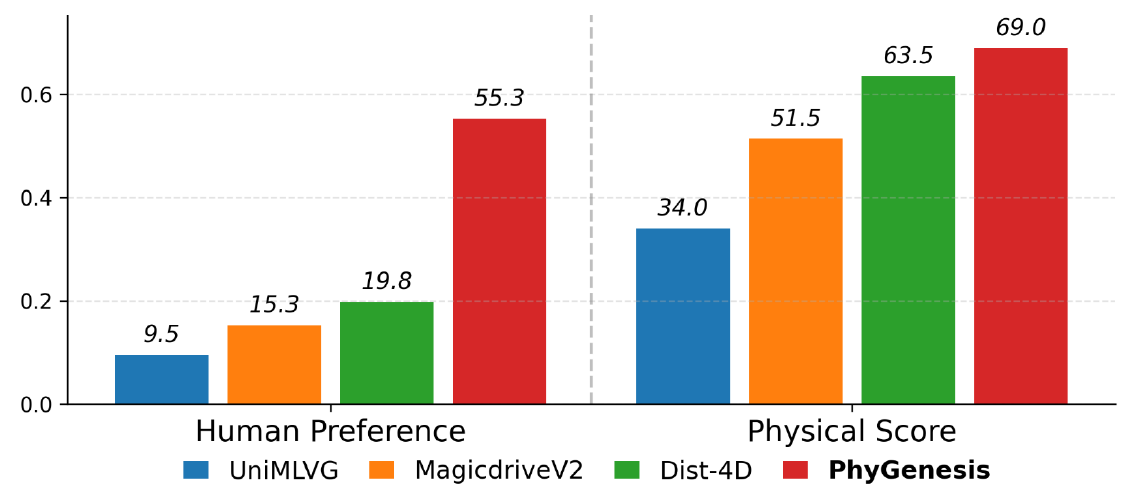} 
\end{minipage}
\hfill
\begin{minipage}[t]{0.49\linewidth}
\centering
\caption{Performance of the physical condition generator for trajectory rectification. We report the 6-DoF L2 distance to ground truth, comparing w/ and w/o the physical condition generator.}
\label{tab:performance_physical_model}
\begin{small}
\begin{sc}
\resizebox{\linewidth}{!}{%
\begin{tabular}{lccc}
\toprule
\textbf{Method} & \textit{nuScenes} & \textit{CARLA Ego} & \textit{CARLA ADV} \\
\midrule
w/o Phys. Cond. Generator & 0.21 & 1.78 & 1.05 \\
W/ Phys. Cond. Generator   & \textbf{0.19} & \textbf{0.65} & \textbf{0.86} \\
\bottomrule
\end{tabular}}
\end{sc}
\end{small}
\end{minipage}

\end{table}

\begin{figure}[tb]
  \centering
  \vskip -0.1in
  \includegraphics[width=\linewidth]{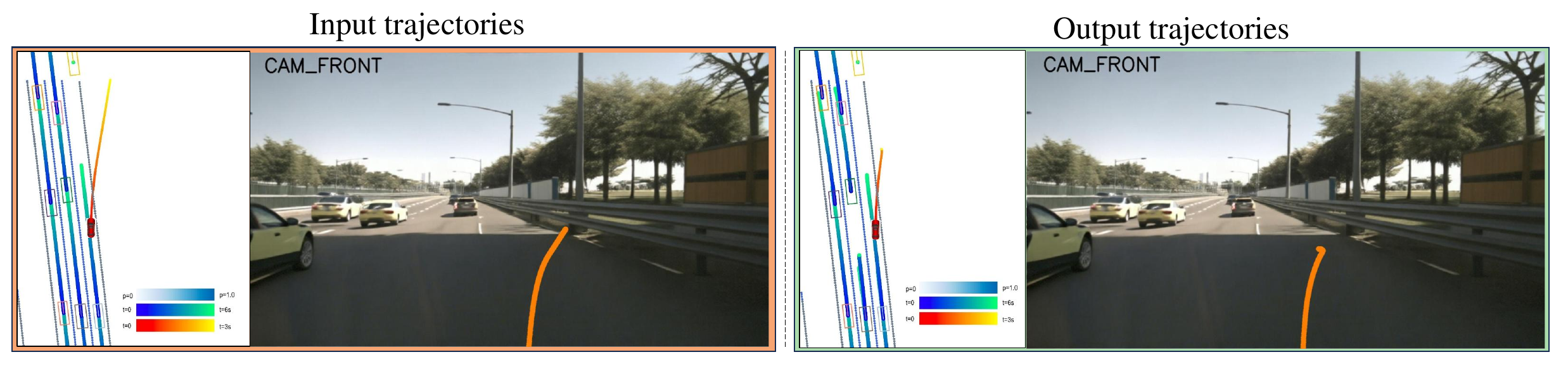}
  \vskip -0.1in
  \caption{
 Performance of the physical condition generator (visualized in BEV and camera views). The input trajectory on the left penetrates the guardrail, while our condition generator rectifies it, resulting in a collision with the guardrail and a stop.
  }
  \label{fig:input_output}
  \vskip -0.1in
\end{figure}

In this section, we evaluate Physical Condition Generator. Following Section~\ref{exp:complete framework}, we construct 2D trajectory conditions on \textit{nuScenes}, \textit{CARLA Ego}, and \textit{CARLA Adv}, and report the 6-DoF L2 distance to ground truth before vs.\ after rectification (on the physical agent and its interaction partner). Table~\ref{tab:performance_physical_model} shows consistent error reductions across datasets, and Figure~\ref{fig:input_output} gives a qualitative example of trajectory rectification. Gains can be observed on \textit{nuScenes} since the input provides only 2D $(x,y)$ and the model mainly recovers the missing 4-DoF. In contrast, for physics-violating trajectories in \textit{CARLA Ego} and \textit{CARLA Adv}, the model achieves substantial error reduction, demonstrating its effectiveness.



\subsection{Ablation Study}
\label{sec:ablation}

\begin{figure}[tb]
  \centering
  \includegraphics[width=\linewidth]{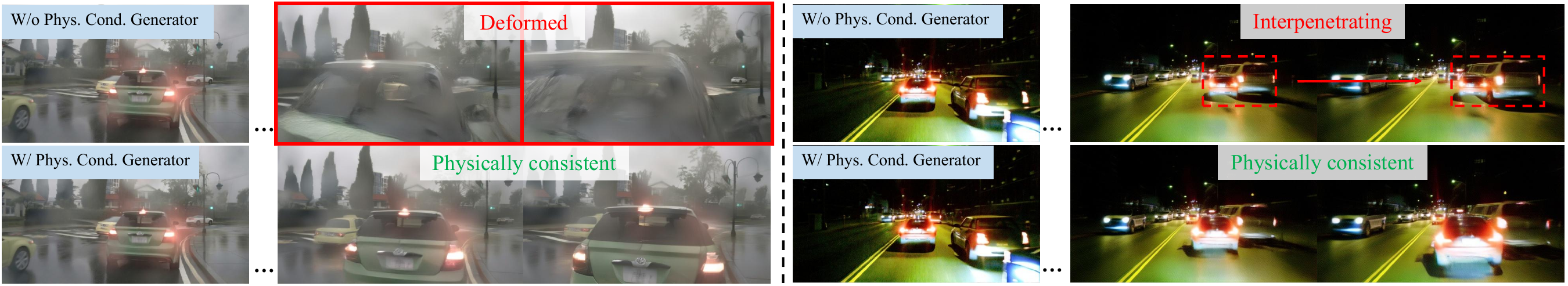}
  \vskip -0.1in
  \caption{
  Effect of the Physical Condition Generator under physics-violating trajectories: it reduces penetration artifacts between vehicles and the environment.
  }
  \vskip -0.05in
  \label{fig:with_physical_model_v2}
\end{figure}

\begin{figure}[tb]
  \centering
  \includegraphics[width=\linewidth]{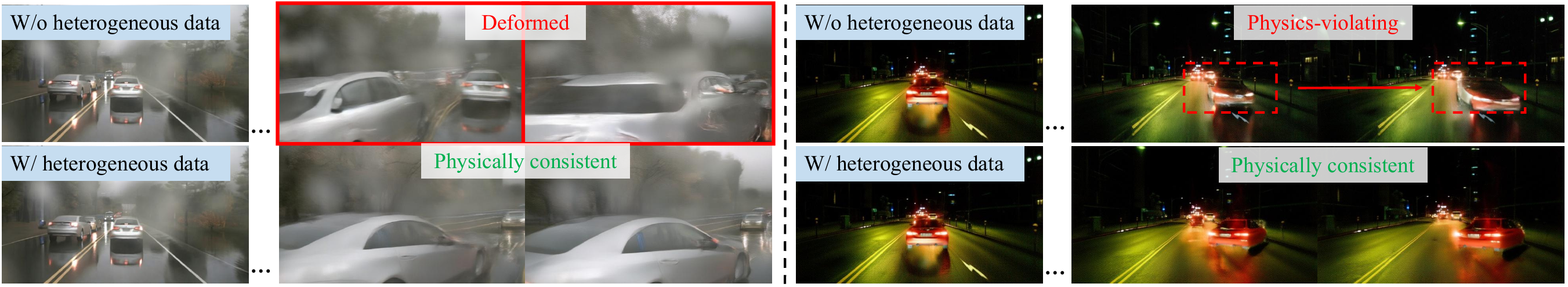}
  \vskip -0.04in
  \caption{
Effect of heterogeneous co-training: it improves the generation quality in physically challenging frames.
  }
  \label{fig:with_physical_data}
\end{figure}


\begin{table*}[tb]
\centering
\vskip -0.1in
\caption{Ablation study. ``Mixed Data'' indicates co-training our heterogeneous dataset, and ``Phy-Model'' denotes the integration of our Physical Condition generator. The best results are highlighted in \textbf{bold}.}
\vskip -0.1in
\label{tab:ablation_study}
\resizebox{0.95\textwidth}{!}{%
\begin{tabular}{cc cccc cccc cccc}
\toprule
\multicolumn{2}{c}{\textbf{Components}} & \multicolumn{4}{c}{\textit{nuScenes}} & \multicolumn{4}{c}{\textit{CARLA Ego}} & \multicolumn{4}{c}{\textit{CARLA ADV}} \\
\cmidrule(lr){1-2} \cmidrule(lr){3-6} \cmidrule(lr){7-10} \cmidrule(lr){11-14}
Mixed Data & Phy-Model & FID $\downarrow$ & FVD $\downarrow$ & PHY $\uparrow$ & Pref. $\uparrow$ & FID $\downarrow$ & FVD $\downarrow$ & PHY $\uparrow$ & Pref. $\uparrow$ & FID $\downarrow$ & FVD $\downarrow$ & PHY $\uparrow$ & Pref. $\uparrow$ \\
\midrule
\checkmark & \checkmark & \textbf{10.24} & \textbf{40.41} & \textbf{0.97} & 0.36 & \textbf{11.03} & \textbf{72.48} & \textbf{0.71} &\textbf{0.55} & 9.28 & \textbf{77.83} & \textbf{0.87} & \textbf{0.57}\\
\checkmark & \(\times\)   & 10.70 & 44.43 & 0.97 & 0.26& 11.81 & 116.51 & 0.65 & 0.19 & \textbf{9.18} & 89.25 & 0.85 & 0.28\\
\(\times\) & \checkmark   & 10.53 & 41.06 & 0.97 & \textbf{0.38}& 12.74 & 82.39 & 0.71 & 0.27 & 9.83 & 89.83 & 0.84 & 0.15\\
\bottomrule
\end{tabular}%
}
\end{table*}

In this section, we ablate the two key components of \textbf{PhyGenesis}: the \textit{Physical Condition Generator} and the \textit{heterogeneous co-training} with physics-rich CARLA data (Mixed Data). Table~\ref{tab:ablation_study} summarizes the results. 

\noindent\textbf{Effect of the Physical Condition Generator.}
Rows (1) and (2) in Table~\ref{tab:ablation_study} show that the physical condition generator improves both visual quality and physical consistency, with the largest gains on the physically challenging \textit{CARLA Ego} and \textit{CARLA Adv} sets. Figure~\ref{fig:with_physical_model_v2} further illustrates it mitigates penetration and deformation artifacts, yielding more physically consistent interactions among vehicles and nearby structures under physics-violating trajectory conditions.

\noindent\textbf{Effect of heterogeneous co-training.}
Rows (1) and (3) in Table~\ref{tab:ablation_study} shows that incorporating physics-rich CARLA data consistently improves performance on the physically challenging data. On \textit{CARLA ADV}, heterogeneous training reduces FVD from 89.83 to 77.83 ($\sim$13.4\% relative), along with a substantial gain in preference (0.13 to 0.53). Figure~\ref{fig:with_physical_data} shows that training on nominal \textit{nuScenes} alone often deforms vehicles in challenging interactions, while heterogeneous co-training produces sharper, more physically coherent dynamics.

\section{Conclusion}
\label{sec:conclusion}

We present PhyGenesis, a novel driving world model for physically consistent and high-fidelity multi-view video generation. By explicitly handling trajectory feasibility and physics-enhanced video generation, our approach outperforms existing methods in both visual fidelity and physical consistency, particularly under challenging trajectory conditions.
Our framework enables more reliable simulation of safety-critical events such as collisions and off-road behaviors. By better aligning planner- or simulator-provided trajectory conditions with physically consistent visual world modeling, PhyGenesis offers a practical building block toward simulation-driven evaluation and safety testing in autonomous driving.




%
%
\bibliographystyle{splncs04}
\bibliography{main}

\newpage

\appendix 

\section{User Study Setting}
\begin{figure}[htbp]
  \centering
  \centerline{\includegraphics[height=0.7\textheight]{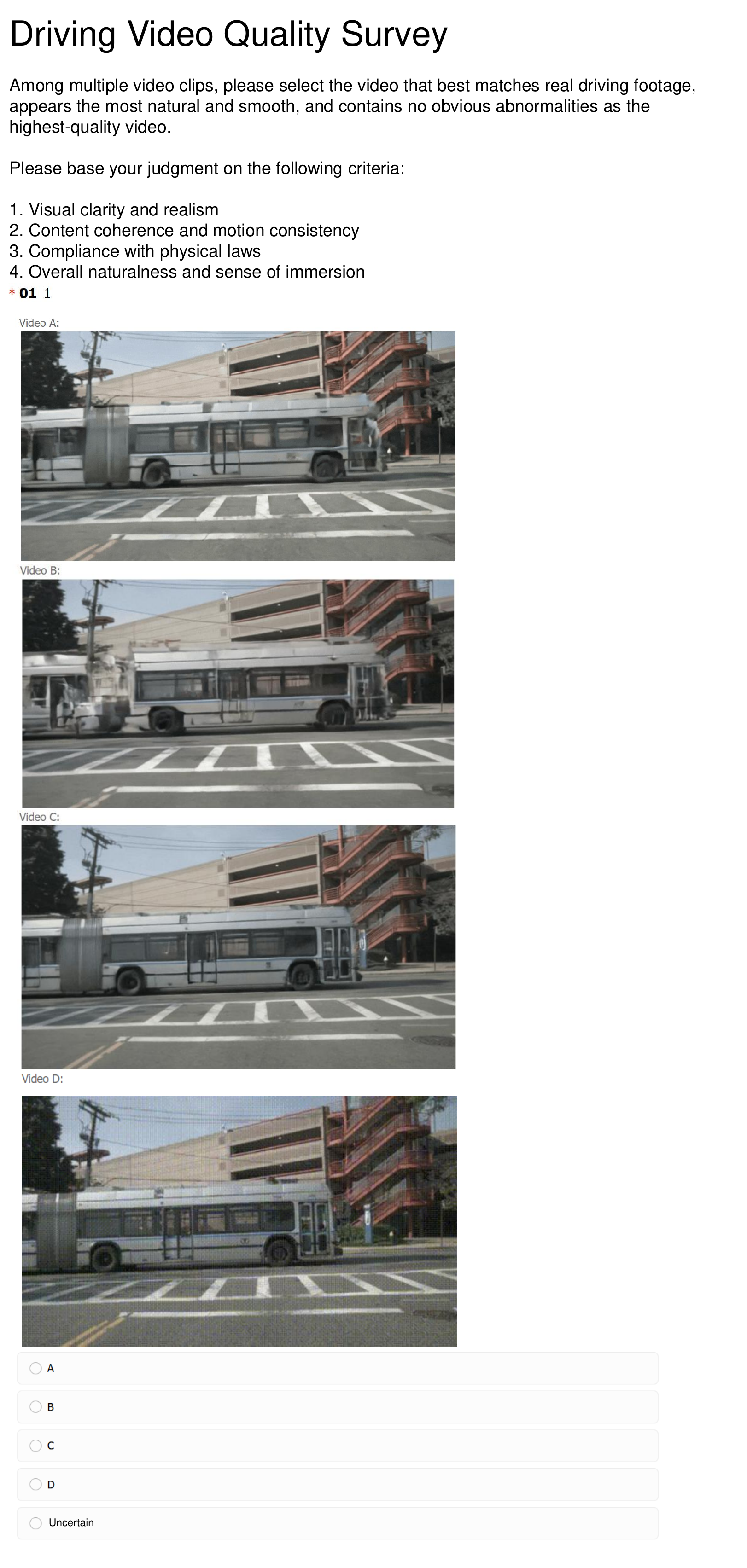}}
  \caption{The questionnaire used to evaluate the visual quality of videos generated by different baselines in the user study.}
  \label{human}
\end{figure}

In our experiments, participants are shown videos generated by different baselines under the same conditions and are asked to select the one with the highest visual quality. In addition to selecting one of the candidate videos, participants may also choose an ``uncertain'' option. A selected video is assigned 1 point, while for an ``uncertain'' response, the score is distributed equally among all compared videos, with each video receiving $1/N$ point, where $N$ is the number of videos in the comparison. The final score is computed as the total points received divided by the total number of comparisons. The questionnaire used in our study is shown in Figure~\ref{human}.

The user studies reported in Table~1, Figure~7, and Table~4 of the main paper are conducted separately. For each study, we randomly sample generated videos from five scenes in the corresponding evaluation dataset for comparison. Each human preference score is based on 150 responses collected from 30 participants. In total, we collect 1,050 responses across all user studies.

\section{Style Transfer Model Used for Video Generation}

\begin{figure}[ht]
  \centering
  \centerline{\includegraphics[width=0.9\columnwidth]{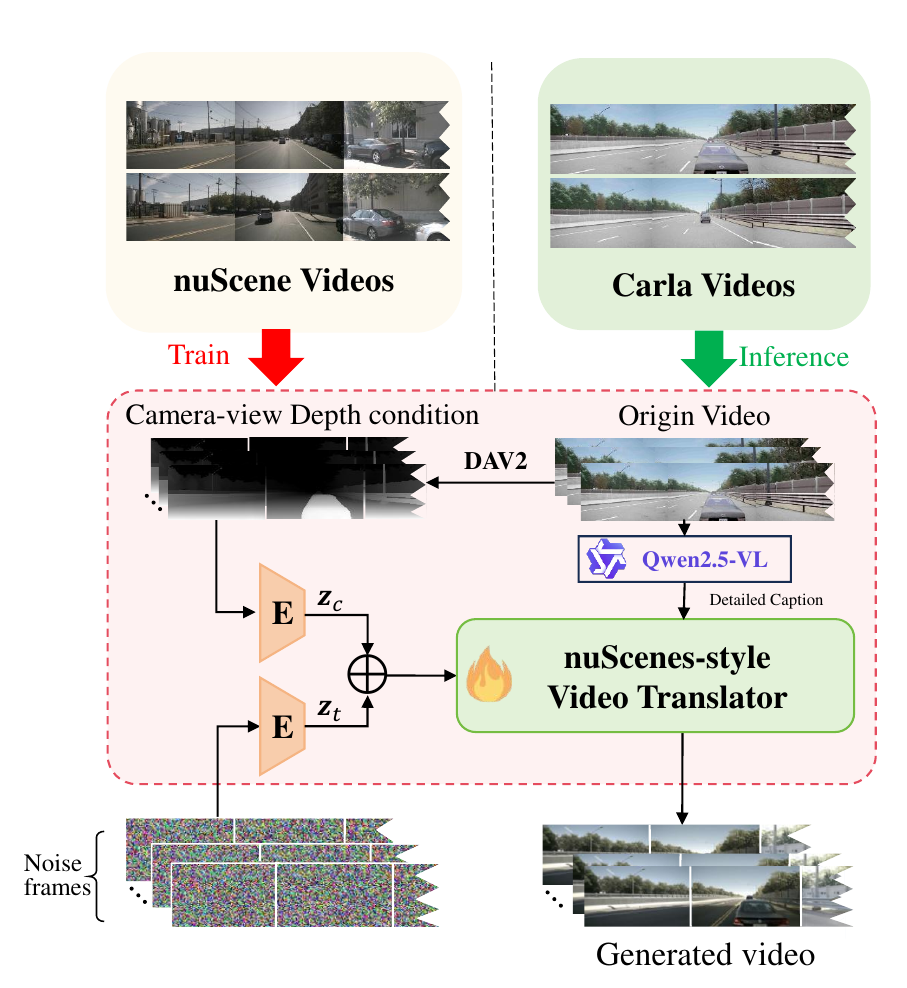}}
  \caption{Architecture of the style transfer model used to translate CARLA videos into the \textit{nuScenes} visual style for fair comparison.}
  \label{transferModel}
\end{figure}

\begin{figure}[ht]
  \centering
  \centerline{\includegraphics[width=\columnwidth]{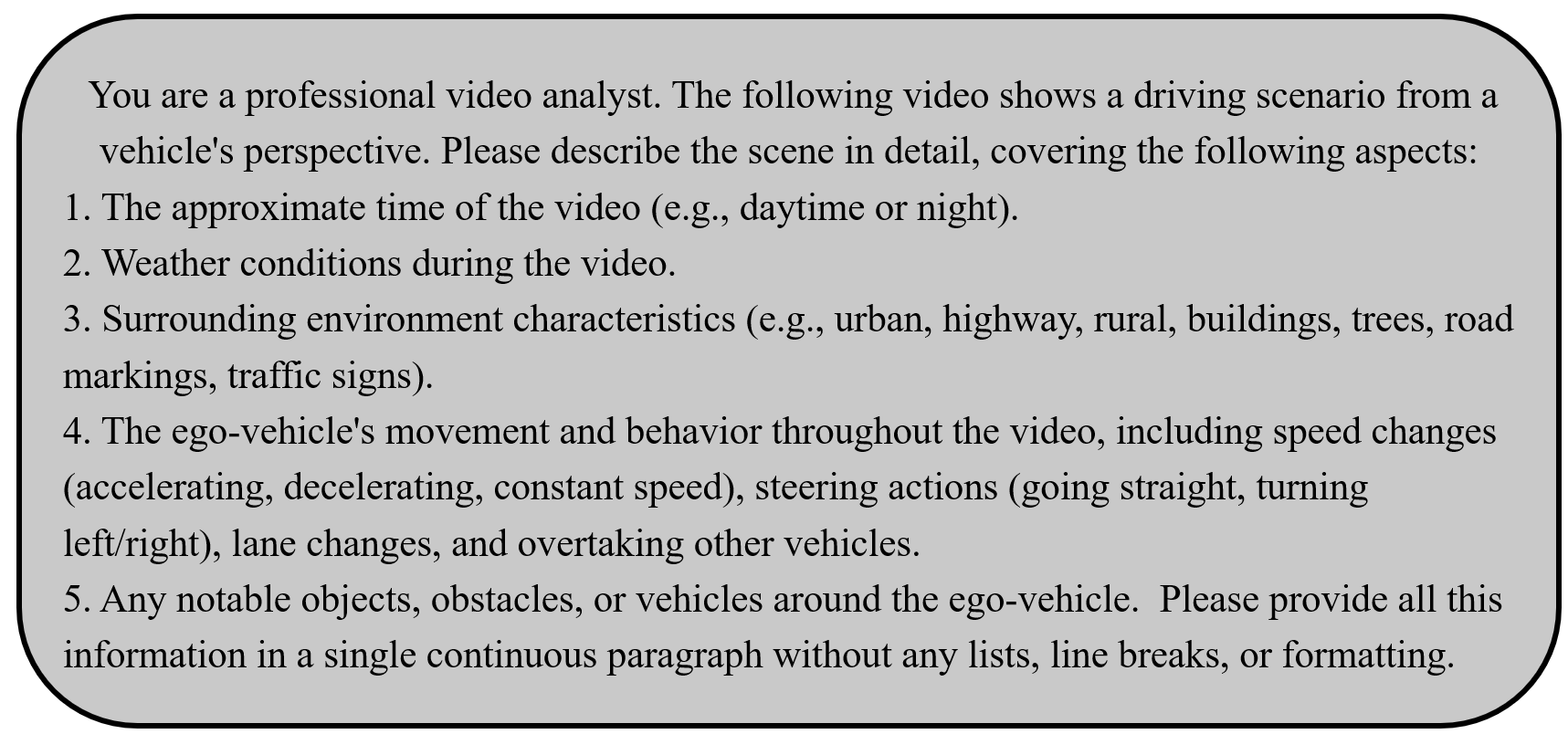}}
  \caption{Prompt used in the VLM for generating detailed video-level captions.}
  \label{transferPrompt}
\end{figure}

Since the compared baselines are primarily trained on \textit{nuScenes}, we introduce a style transfer model to ensure a fair comparison. Our transfer model is built upon Wan2.1-Fun-V1.1-1.3B-Control \cite{wan2025wan} and follows the same rectified-flow video generation framework. Specifically, we adapt the generation model to the style transfer setting by conditioning it on per-frame depth information and detailed textual descriptions, so that CARLA clips can be translated into the \textit{nuScenes} visual style. After stylizing the full CARLA videos into the \textit{nuScenes} domain, we can provide all compared video
generation methods with \textit{nuScenes}-style initial frames on the CARLA benchmark, and also obtain stylized videos to be the groundtruth for computing FID and FVD.

The architecture of the transfer model is shown in Figure~\ref{transferModel}. Specifically, we use Depth Anything V2 (DAV2) \cite{yang2024depth} to extract per-frame depth maps from the input video, and, similar to the main model, concatenate the resulting depth latents with noisy video latents as generation conditions, without using the initial frame as a conditioning input. In addition, we use Qwen2.5-VL \cite{qwen2.5-VL} to produce a detailed description for the entire video, as the semantic content of the video is already fixed. The prompt used for generating video-level captions is shown in Figure~\ref{transferPrompt}.

Importantly, the transfer model is trained entirely on \textit{nuScenes} data, rather than on heterogeneous mixed-domain data. This design encourages the model to learn the appearance characteristics of the \textit{nuScenes} domain more faithfully. Since the initial frame is not used as a conditioning input, the model’s generation inherently reflects the \textit{nuScenes} style rather than the specific appearance of the input video. During inference, the trained transfer model is applied to CARLA videos to translate them into the \textit{nuScenes} style.

\setlength{\parskip}{1pt}
\noindent \textbf{Training Objective.} Following the rectified-flow formulation used in the main paper, let $\mathbf{z}_1$ denote the clean target video latent, let $\mathbf{z}_0 \sim \mathcal{N}(\mathbf{0}, \mathbf{I})$ be a random noise latent, and let $t \in [0,1]$ be the sampled timestep. The noisy latent is constructed as
\begin{equation}
    \mathbf{z}_t = t\mathbf{z}_1 + (1-t)\mathbf{z}_0 .
\end{equation}
The transfer model is trained with the same flow-matching objective, conditioned on the video-level caption $\mathbf{c}_{text}$, and the per-frame depth condition $\mathbf{c}_{depth}$:
\begin{equation}
    \mathcal{L}_{\text{transfer}}=
    \mathbb{E}_{\mathbf{z}_0,\mathbf{z}_1,t}
    \left\|
    u_\theta(\mathbf{z}_t,t,\mathbf{c}_{text},\mathbf{c}_{depth})
    -
    (\mathbf{z}_1-\mathbf{z}_0)
    \right\|_2^2 .
\end{equation}

\section{Implementation Details of Different Baselines}

In this paper, we compare our method with three baselines, all evaluated using their official pretrained weights. For all methods, we generate videos consisting of 33 frames. Our method, MagicDriveV2 \cite{gao2025magicdrive}, and DiST-4D \cite{guo2025dist} generate the entire video in a single pass. In contrast, UniMLVG \cite{chen2025unimlvg} can generate at most 19 frames at a time. Therefore, we follow its official autoregressive inference code to produce a 33-frame video in two stages: the first pass generates frames 1--14, and the second pass generates frames 14--33, using the intermediate frame as the conditioning frame for the second stage.

Our method generates videos at a resolution of $448 \times 800$. MagicDriveV2 and DiST-4D generate videos at $424 \times 800$, while UniMLVG generates videos at $256 \times 448$, since it is trained only at this resolution.

Our method, MagicDriveV2, and UniMLVG all use the initial RGB frame as the generation condition. DiST-4D additionally uses the depth map of the initial frame as an extra condition, which we obtain using its official preprocessing pipeline. For the CARLA stylized domain, we apply a one-epoch adaptation to the Physical Condition Generator on stylized initial frames from the CARLA training set, while keeping the  Physics-enhanced Multi-view Video Generator unchanged.

\section{Physical-Challenging Scenario Construction in CARLA Ego and CARLA Adv}

We construct physically challenging scenarios in CARLA \cite{Dosovitskiy2017carla} based on the Bench2Drive \cite{jia2024bench2drive} routing setup. Each rollout starts from a valid predefined route while preserving the original map layout, traffic participants, weather, and background behaviors. On top of this standard setup, we perturb the behavior of one designated vehicle to induce physically challenging events. The perturbation is parameterized by two variables: a lateral route offset and a target speed. Specifically, the target speed is sampled from predefined values between 0 and 30 m/s, with uniform random perturbations within the intervals around each value, and the lateral offset is sampled from predefined values between -200 and 200 meters, with uniform random perturbations within the intervals around each value. At scene initialization, we further choose one of three perturbation modes with equal probability: (i) zero lateral offset with randomized target speed, (ii) fixed target speed of 10 m/s with randomized lateral offset, and (iii) both target speed and lateral offset randomized. This design yields a broad spectrum of behaviors ranging from mild deviations to aggressive maneuvers.

For \textit{CARLA Ego}, the perturbed vehicle is the ego vehicle itself. The ego vehicle first follows the default autopilot for a warm-up period of 24 steps. Since our simulator runs at 12 Hz, this corresponds to 2 seconds. After the warm-up, we replace the nominal route with a smoothed and supersampled route, and then apply a cosine-smoothed lateral offset to the future route points so that the vehicle gradually transitions from its current route to the perturbed one. The ego vehicle is then controlled to track this modified route under the sampled target speed. For \textit{CARLA Adv}, we instead perturb a nearby non-ego vehicle. We first spawn an adversarial vehicle near the ego vehicle by sampling a valid waypoint within a local neighborhood of the ego route, including forward, backward, and adjacent-lane candidates within approximately 15 meters. The adversarial vehicle also follows the default controller for the same 24-step warm-up period. Afterward, we transform the ego route into the adversarial vehicle's local initialization frame and apply the same perturbation mechanism. As a result, the physically challenging behavior is centered on a nearby non-ego agent rather than the ego vehicle.

During rollout, we monitor collisions and off-road departures for the perturbed vehicle. Collisions are detected using CARLA collision sensors, and the collided object category is recorded, including vehicles, pedestrians, parked vehicles, traffic objects, and static objects. Off-road events are detected by querying the local lane type and checking whether the vehicle enters sidewalk or shoulder regions. Once an event is triggered, we continue collecting data for 48 additional frames, corresponding to 4 seconds at 12 Hz, and then terminate the rollout. We also terminate the rollout if no event is observed within 120 post-warm-up steps, i.e., 10 seconds. In this way, each collected sequence contains the pre-event context, the event itself, and its short-term consequence. The resulting dataset therefore consists of route-grounded but physically challenging trajectories. \textit{CARLA Ego} emphasizes failures and abrupt maneuvers induced on the ego vehicle, while \textit{CARLA Adv} emphasizes interactions caused by a nearby perturbed non-ego vehicle.


\section{Weighting Design and Ablation Study of $\lambda_{\text{event}}$ and $\lambda_{\text{agent}}$}

In the Physical Condition Generator optimization, $W_{i,t}$ is defined using two scalar hyperparameters to prioritize physically important samples. The temporal weight $\lambda_{\text{event}}$ reweights the loss along the time dimension by assigning larger weights to timesteps near critical events such as collision or off-road. For each event timestep $t_e$, we define a forward window $[s_e, e_e]$, where $s_e = \max(0, t_e - 1)$ and $e_e = \min(T - 1, t_e + 10)$, with $T$ denoting the prediction horizon. Within this window, the temporal weight is initialized at $\lambda_{\text{event}}$ and exponentially decays to $1$:
\[
w_e(t)=
\lambda_{\text{event}}
\exp\!\left(
\frac{\log(1/\lambda_{\text{event}})}{e_e-s_e}(t-s_e)
\right), \qquad t\in[s_e,e_e].
\]
If multiple event windows overlap, we take the maximum weight at each timestep:
\[
W_t^{\text{event}}=\max_e w_e(t),
\]
and set $W_t^{\text{event}}=1$ for timesteps outside all event windows. This design makes the supervision strongest near the onset of a physical event while preserving standard supervision on the remaining horizon.

The agent weight $\lambda_{\text{agent}}$ further reweights the loss along the agent dimension by assigning a larger multiplicative factor to the agents that are most relevant to the physical event. The selection is determined based on the event annotations in the dataset. When the event corresponds to a collision with a static object, such as a fence or a lamp post, only the ego vehicle (or the Adv vehicle) is emphasized. When the event corresponds to a collision with a dynamic participant, such as another vehicle or a pedestrian, we additionally emphasize the agent that is nearest to the ego vehicle (or the Adv vehicle) at the collision timestep. In this way, the optimization focuses on the participants that are most directly related to the event, which helps better model the trajectory behavior around the collision moment.

The ablation study presented in Tables~\ref{tab:performance_event} and~\ref{tab:performance_agent} indicates that different values of $\lambda_{\text{event}}$ and $\lambda_{\text{agent}}$ have limited effect on performance. This suggests that the model is largely robust to these hyperparameters while the weighting mechanism still provides focused supervision around critical events and agents.

\begin{table}[ht]
\centering
\caption{Ablation study on $\lambda_{\text{event}}$.}
\vspace{5pt}
\label{tab:performance_event}
\begin{tabular}{lccc}
\toprule
\textbf{Configuration} & \textit{nuScenes} & \textit{CARLA Ego} & \textit{CARLA ADV} \\
\midrule
baseline & 0.21 & 1.78 & 1.05 \\
$\lambda_{\text{event}}=1$  & 0.17 & 0.56 & 0.69 \\
$\lambda_{\text{event}}=5$  & 0.16 & 0.58 & 0.76 \\
$\lambda_{\text{event}}=10$ & 0.16 & 0.57 & 0.77 \\
\bottomrule
\end{tabular}
\end{table}

\begin{table}[h]
\centering
\caption{Ablation study on $\lambda_{\text{agent}}$.}
\vspace{5pt}
\label{tab:performance_agent}
\begin{tabular}{lccc}
\toprule
\textbf{Configuration} & \textit{nuScenes} & \textit{CARLA Ego} & \textit{CARLA ADV} \\
\midrule
baseline & 0.21 & 1.78 & 1.05 \\
$\lambda_{\text{agent}}=1$   & 0.13 & 0.53 & 0.70 \\
$\lambda_{\text{agent}}=10$  & 0.16 & 0.58 & 0.76 \\
$\lambda_{\text{agent}}=20$  & 0.16 & 0.62 & 0.90 \\
\bottomrule
\end{tabular}
\end{table}

\end{document}